\pdfoutput=1

\documentclass[10pt, a4paper]{article}

\usepackage{lrec2026} 
\usepackage{lineno}

\usepackage{times}
\usepackage{latexsym}
\usepackage{tabularx}
\usepackage{adjustbox}
\usepackage{booktabs}
\usepackage{graphicx}
\usepackage{natbib}
\usepackage{makecell}
\usepackage{rotating}
\usepackage{lscape}
\usepackage{float}
\usepackage{footmisc}

\usepackage[T1]{fontenc}

\usepackage[utf8]{inputenc}

\usepackage{microtype}

\usepackage{inconsolata}

\usepackage{tikz}
\usepackage[most]{tcolorbox}

\usetikzlibrary{shapes.geometric, arrows, positioning}

\tikzstyle{startstop} = [rectangle, rounded corners, minimum width=3cm, minimum height=1cm,text centered, draw=black, fill=red!30]
\tikzstyle{process} = [rectangle, minimum width=3cm, minimum height=1cm, text centered, draw=black, fill=blue!20]
\tikzstyle{decision} = [diamond, minimum width=3cm, minimum height=1cm, text centered, draw=black, fill=green!20]
\tikzstyle{arrow} = [thick,->,>=stealth]

\newtcolorbox{resultbox}{
  colback=gray!10,       
  colframe=gray!10,      
  boxrule=0pt,           
  borderline west={2pt}{0pt}{gray!80!black},  
  enhanced,
  breakable,
  boxsep=0pt,
  left=4pt, right=2pt,   
  top=2pt, bottom=2pt
}

\title{Mental Health Disorder Detection Beyond Social Media: \\ A Systematic Review of Available Datasets}

\name{Sadiya Sayara Chowdhury Puspo\textsuperscript{1}, Ana-Maria Bucur\textsuperscript{2}, Stevie Chancellor\textsuperscript{3},\\
\large\textbf{Özlem Uzuner\textsuperscript{1}, Marcos Zampieri\textsuperscript{1}}\vspace{2mm}} 

\address{\textsuperscript{1}George Mason University, USA \\ \textsuperscript{2}Università della Svizzera italiana, Switzerland \\
\textsuperscript{3}University of Minnesota, USA\\
}

\abstract{
Detecting mental health disorders in a timely manner is an important societal challenge. NLP and machine learning (ML) methods used to assist with detection rely on data collected primarily from social media. However, such datasets often have sampling biases and inherent ethical and privacy issues. One avenue to overcome these limitations is non-social media data. We present the first comprehensive review of non-social media, free-text datasets for mental health research. We use the PRISMA methodology to conduct our survey and we review datasets available in multiple languages. We find that non-social media free-text based datasets are predominantly focused on English and on detecting depression. These datasets also vary in demographics, platforms, data types, annotation techniques, and methodologies. This systematic review also reveals key gaps and highlights opportunities to develop more diverse, reliable and clinically-relevant resources.
 \\ \newline  }

\begin{document}

\maketitleabstract

\section{Introduction}
\label{sec:intro}

The prevalence of mental health disorders is a global concern. In the USA, for example, one in every four adults experiences a diagnosable mental health disorder each year\footnote{\scriptsize\href{https://www.hopkinsmedicine.org/health/wellness-and-prevention/mental-health-disorder-statistics}{https://www.hopkinsmedicine.org/health/wellness-and-prevention/mental-health-disorder-statistics}}. Furthermore, research shows that the majority of individuals who die by suicide have an identifiable mental health condition such as depression \cite{bailey2011suicide} or substance use disorder \cite{lynch2020substance}.


The limited access to mental health care services has become an urgent societal issue \cite{cummings2013improving,hester2017lack}. This motivated research on applying NLP and ML models to social media data for early identification of mental disorders and supporting individuals at risk. Social media platforms such as Twitter (\citealp{chatterjee2022suicide}; \citealp{khafaga2023deep}), Reddit (\citealp{boettcher2021studies}; \citealp{adil2022corpus}) and Facebook (\citealp{calvo2017natural}; \citealp{islam2018depression}) offer a rich, unobtrusive stream of user-generated content that captures real-life expressions with reduced reporting bias, aiding in the detection and understanding of mental disorders \cite{sametouglu2024value,raihan2024mentalhelp,raihan2026large}. 

Using social media data for mental health screening raises ethical and privacy concerns regarding user consent, construct validity, and the potential for algorithmic misuse \cite{chancellor2019human,nicholas2020ethics,chancellor2020methods}. Collecting data from social media can lead to unintended biases as the data may only reflect the experiences of individuals who are willing to openly discuss their mental health online, which primarily includes those who are active on social media platforms \cite{chancellor2019taxonomy}. Finally, demographics vary across platforms and they may not be representative of the general population. For instance, X users are primarily male, while TikTok and Instagram users are mostly female; likewise, many platforms tend to be used by teens and young adults \cite{olteanu2019social,zhao2022biases}. Pragmatically, many social media sites have limited access to APIs for research.

The widespread use of social media data in this domain is partly due to the significant shortage of non-social media free-text based datasets related to mental health disorders derived from clinical and other reliable sources. Clinical data, such as electronic medical records (EMRs) or electronic health records (EHRs) with discharge summaries or clinical notes, psychiatric interview transcripts, and responses to standardized open-ended questionnaires, offer rich, detailed insights into patients’ mental health. Unlike social media data, these sources contain carefully documented information from healthcare professionals, including diagnostic details, symptom descriptions, and treatment histories, often supported by validated clinical scales.

Despite their great potential, non-social media free-text based datasets remain underexplored primarily due to privacy concerns, data access challenges, and annotation complexities. This scarcity presents a barrier to advancing robust, generalizable NLP models that can be effectively integrated into clinical practice. We aim to address this gap by systematically reviewing non-social media free-text based datasets for mental health research. We explore their diversity in terms of source, structure, clinical annotation, types, and research adoption to guide future efforts in dataset development and application. Previous related surveys have primarily focused on datasets from social media platforms \cite{harrigian2021state,rissola2020beyond,skaik2020using,abdulsalam2024suicidal,bucur2025datasets,bucur2025state}. To the best of our knowledge, this is the first systematic review\footnote{\scriptsize\href{https://github.com/SadiyaPuspo/MHD-Beyond-Social-Media-Datasets-Review}{https://github.com/SadiyaPuspo/MHD-Beyond-Social-Media-Datasets-Review}} \cite{grant2009typology} of free-text based mental health data sources beyond social media. 

We address the following research questions:

\vspace{-2mm}

\paragraph{RQ\textsubscript{1}:} What non-social media free-text based datasets are available for mental health research, and how do they vary by source, structure, and population? 


\vspace{-2mm}

\paragraph{RQ\textsubscript{2}:} How are mental health conditions defined and labeled in these datasets, and what are the clinical implications of these labeling methods?


\vspace{-2mm}

\paragraph{RQ\textsubscript{3}:} What factors contribute to the popularity and adoption of non-social media free-text based mental health datasets in research?


\section{Methods}

\begin{figure}
\centering
\includegraphics[width=0.8\columnwidth]{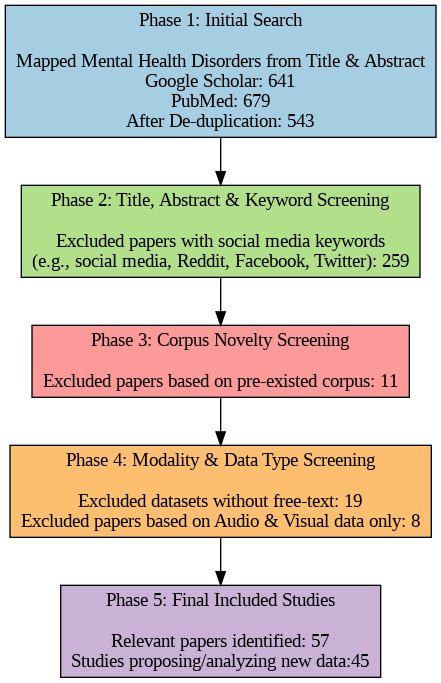}
\caption{PRISMA flowchart illustrating the systematic literature search and screening process.}
\label{fig:prisma}
\end{figure}

This systematic review adopts the PRISMA\footnote{\scriptsize\href{https://www.prisma-statement.org/prisma-2020}{https://www.prisma-statement.org/prisma-2020}} reporting guidelines to comprehensively map and systematically analyze the landscape of the free-text based datasets in mental health research beyond social media, building on prior frameworks (e.g. \citet{templier2015framework, zarate2022exploring, pazdur2025risk}). PRISMA is a standardized guideline aimed at ensuring clear and thorough reporting of systematic reviews and meta-analyses. It features a 27-item checklist that helps authors include all essential components of their review, from initial identification to final conclusions. A key part of PRISMA is its flow diagram (Figure \ref{fig:prisma}), which visually maps the process of selecting studies, making the review process more transparent and easy to follow \cite{page2021prisma}.

To follow the checklists of PRISMA, we began with a systematic literature search that continued through June 2025. We used Publish or Perish\footnote{\scriptsize\href{https://harzing.com/resources/publish-or-perish}{https://harzing.com/resources/publish-or-perish}} to query multiple academic databases, including Google Scholar, and PubMed. The searches were carried out separately for key mental health related terms: `depression', `anxiety', `suicidal ideation', `suicide', `mental disorder', `mental health crisis', and `mental health disorder', combined with targeted keywords `identification', `detection', `prediction', and `analysis' to capture each relevant disorder type of mental health and associated dataset corpus. We then manually screened the abstracts and included studies that explicitly referenced the use of any datasets or the application of any computational models or techniques after de-duplication, resulting in 543 papers within our initial scope. The rationale for selecting depression and anxiety as key terms is that they are among the most prevalent mental health disorders worldwide \cite{GBD2021_IHME_2024}. Additionally, given that suicide remains a leading cause of death globally among adults \cite{CDC_SuicidalThoughts_2025}, we prioritize the inclusion of these mental health conditions in our study.

To ensure focus on non-social media sources, we excluded papers whose titles or abstracts included keywords related to social media platforms or mentions of datasets scraped from them, such as `social media', `Reddit', `Facebook', or `Twitter'. This filtering was essential to isolate papers based on clinical notes, semi-structured interviews, and discharge summaries from other non-social media platforms, and we ended up with 259 total papers after this filtering. Subsequently, we conducted corpus novelty screening to remove studies based on pre-existing datasets, followed by modality screening to exclude papers that focused solely on audio or visual data, and data type screening to filter out datasets containing only numerical or categorical values, such as scale-based survey responses. Through this multi-phase screening, we identified 57 relevant papers, of which 40 proposed or analyzed newly collected free-text based mental health datasets.

In addition to keyword-based search, we followed a backtracking strategy; whenever a paper used an existing dataset, we traced it back to the original dataset publication, even if it did not contain our predefined keywords. We also examined all papers cited in the literature review or background sections of included studies and explored any dataset-related papers mentioned in relevant survey papers found from the initial search. As a result, our final collection may include important dataset papers whose titles or abstracts do not directly match the initial search terms, but were directly aligned with our objectives. This process yielded 5 additional papers, for a total of 45.

Several datasets were initially annotated for mental disorders, but their final task focused on emotion classification, including loneliness, fear, anger, hopelessness, and self-identification. Since emotions like loneliness and hopelessness can serve as indicators of depression \cite{rholes1985relationship} or suicidal ideation \cite{baryshnikov2020role}, we decided to include these studies in our survey.

\section{Distribution Analysis} 

In this section, we provide information on the 45 free-text based datasets included in this systematic review. Table \ref{tab:mental_health_datasets} provides a comprehensive summary of the final set of datasets. The datasets were published between 2004 and 2025 in multiple languages, including English, Chinese, Polish, Korean, Japanese, Arabic, and some code-mixed. These datasets cover a broad spectrum of mental health conditions, such as depression, postnatal depression, anxiety, schizophrenia, suicidal ideation, Post-Traumatic Stress Disorder, and bipolar disorder, collected from platforms like clinics, colleges, mobile apps, and therapy sessions. Data types include interview transcripts, essays, discharge summaries, clinical records, forum posts, and suicide notes. While some datasets are publicly available, others are restricted or require agreements, with a few lacking availability details. We show their distribution below across disorders, languages, platforms, data types, demographics, and availability types.

\begin{table*}[!h]
\centering
\scalebox{0.47}{
\begin{tabular}{lcccccccccc}
\toprule
\textbf{Dataset} & \textbf{Language} & \makecell{\textbf{Mental} \\ \textbf{Disorder}} & \textbf{Platform} & \makecell{\textbf{Data} \\ \textbf{Type}} & \makecell{\textbf{Annotation} \\ \textbf{Procedure}} & \makecell{\textbf{Annotation} \\ \textbf{Instrument}} & \textbf{Label} & \textbf{Size} & \textbf{Availability} & \textbf{Citation} \\
\midrule

\citet{pestian2012sentiment} & EN & SD & CLINIC & SD Notes & Manual & Krippendorff's~$\alpha$
 & 15 & 1004 & DUA & 236 \\
\citet{uzuner2017natural} & EN & DP, PTSD, OCD, BD & CLINIC & Clinical Notes & Manual & - & 4 & 816 & DUA & 45 \\
\citet{jackson2017natural} & EN & SMI & CLINIC & EHR(DS) & Manual & Text Hunter & 50 & 37,211 & RSTR & 233 \\
\citet{low2010detection} & EN & MDD & CLINIC & Interview & Manual & LIFE & 2 & 139 & RSTR & 313\\
\citet{pestian2010suicide}  & EN & SD & CLINIC & SD Notes & Manual & Ontology & 2 & 66 & RSTR & 322\\
\citet{geraci2017applying} & EN & MDD \& DD & CLINIC & EMR & Manual & DSM-IV & 2 & 861 & RSTR & 88\\

\citet{zhou2015identifying}  & EN & DP & CLINIC & EHR (DS) & Manual & - & 3 & 1200 & UNK & 74\\
\citet{hcet2021} & EN & DP & CLINIC & EHR & Manual & ICD-9 \& AD & 2 & 10,148 & UNK & 53 \\
\citet{marrie2018effects} & EN & DP \& AX & CLINIC & Interview & Manual & DSM-IV & - & 308 & UNK & 51 \\
\citet{diederich2007ex} & EN & SZ & CLINIC & Essay & Manual & - & 2 & 56 & UNK & 71 \\
\citet{poulin2014predicting} & EN & SD & CLINIC & Clinical Notes & Manual & - & 3 & 210 & UNK & 233 \\
\citet{saini2014assessment} & EN & \makecell{SZ, DP, BD, \\SD \& Other} & CLINIC & \makecell{Interview \&\\ Clinical Notes} & Manual & - & 2 & 198 & UNK & 26 \\

\citet{milne2016clpsych} & EN & MHD & FORUM & Post & Manual & \makecell{Fleiss’s Kappa \& \\ Cohen's Kappa} & 4 & 1227 & PUB & 142\\
\citet{he2017automated} & EN & PTSD & FORUM & Essay & Manual & DSM-IV \& CAP Scale & 2 & 300 & UNK & 95\\
\citet{tyshchenko2018depression} & EN & DP \& AX & FORUM & Post & Self-disclosure & - & 2 & 16,975 & UNK & 55 \\
\citet{communityVScontrol} & EN & DP, BD \& SD & FORUM & Post & Self-disclosure & - & 2 & 267,964 & UNK & 303\\

\citet{aich2022towards} & EN & SZ \& BD & COLLEGE & Interview & Manual & DSM-V \& DSM-IV & 3 & 644 & DUA & 11\\
\citet{rude2004language} & EN & DP & COLLEGE & Essay & Manual & BDI \& IDD-L & 3 & 124 & UNK & 1543\\
\citet{rheault2016expressions} & EN & AX & DEBATE & Political Speech & Manual & - & 2 & 4000 & PUB & 17 \\
\citet{hull2020two} & EN & DP \& AX & Telemedicine Platform & Message & Manual & \makecell{PHQ-9 \\ GAD-7} & \makecell{PHQ-9 \\ GAD-7} & 10,718 & DUA & 43 \\
\citet{krishnamurti2022identification} & EN & PD & APP & Essay & Manual & EPDS & 2 & 1,091 & UNK & 14 \\
\citet{nobles2018identification} & EN & DP \& SD & Phone & SMS & Manual & Self-Disclosure & 2 & 94 & UNK & 127 \\
\citet{howes2014linguistic} & EN & DP \& AX & Online Therapy Chat & Dialogue & Mixed & MALLET \& LIWC & 2 & 882 & UNK & 78\\
\citet{ringeval2019avec} & EN & DP \& PTSD & Virtual Agent & Interview & Manual & PHQ-8 & 5 & 275 & DUA & 402\\
\citet{schoene2016automatic}  & EN & DP \& SD & MIXED  & SD Notes \& Articles & Manual & - & 12 & 426 & UNK & 64\\
\citet{gratch-distress} & EN & DP, AX \& PTSD & MIXED & Interview & Manual & - & - & 621 & DUA & 759\\
\citet{ghosh2020cease}  & EN & DP \& SDI & MIXED & SD Notes \& Book & Manual & Cohen’s Kappa & 15 & 2393 & PUB & 45\\
\citet{tasnim-depac}  & EN & DP \& AX & mTurk & Interview & Manual & \makecell{PHQ-9 \\ GAD-7} & \makecell{PHQ-9 \\ GAD-7} & 2674 & UNK & 29\\
\citet{caicedo2020assessment}  & EN-ES & DP \& SDI & Social Network \& Forum & Post & Manual & Cohen’s Kappa & 4 & 102 & PUB & 24\\
\citet{wu2020using}  & EN-ZH & \makecell{MDD, MinDP, SZ, \\ BD, AjD, DEM \& Other} & CLINIC & EHR(DS) & Manual & \makecell{DSM-IV \\ Sheehan Disability Scale} & 9 & 4,836 & RSTR & 122\\

\hline
\hline

\citet{cmdc} & ZH & MDD & CLINIC & Interview & Manual & HAMD \& PHQ-9 &2 & 78 & PUB & 48\\
\citet{jiang2022mmda}  & ZH & DP \& AX & CLINIC & Interview & Manual & HAMD \& HAMA & 3 & 1025 & PUB & 9\\
\citet{xu2025identifying}  & ZH & DP \& AX & CLINIC & EMR & Manual & DSM-V \& ICD-10 & 4 & 1,160 & PUB & 1\\
\citet{kangning} & ZH & DP & CLINIC & Interview & Manual & MADRS & 2 & 113 & DUA & 5\\
\citet{ive2024data} & ZH & AX & CLINIC & EHR & Manual & ICD-9 \& ICD-10 & 2 & 84,426 & RSTR & 0\\
\citet{li2023detection} & ZH & DP \& SD & CLINIC & Interview & Manual & HAMD & 3 & 305 & UNK & 10 \\
\citet{shen2022automatic} & ZH & DP  & APP & Interview & Manual & SDS & 2 & 162 & PUB & 146\\
\citet{mo2024multimodal} & ZH & AX  & Phn. Recording & Essay & Manual & GAD-7 & 3 & 227 & UNK & 11\\

\hline
\hline
\citet{shin2022detection} & KO & DP, AX \& SD & CLINIC & Interview & Manual & \makecell{PHQ-9, HDRS \\ BAI \& BSS} & 2 & 166 & DUA & 22\\
\citet{figueroa2022automatic}  & KO & SZ \& FEP & CLINIC & Interview & Manual & DSM-V, PANSS & 3 & 133 & RSTR & 33\\
\citet{wawer2022single}  & PL & SZ & CLINIC & Interview & Manual & ICD-10 & 2 & 94 & UNK & 20\\
\citet{oh2024development}  & ES-CL & DP & CLINIC & Interview & Manual & DSM-V & 8 & 451 & UNK & 2\\
\citet{hamalainen2021detecting}  & TH & DP & FORUM & Post & Manual & Key-word Search & 2 & 944 & PUB & 18\\
\citet{hiraga2017predicting} & JA & DP & FORUM & Post & Self-disclosure & - & 2 & 108 & UNK & 29\\
\citet{alghamdi2020predicting}  & AR & DP & FORUM & Post & \makecell{Self-disclosure \& \\ Manual} & \makecell{DSM-5, PHQ-9 \\ QIDS-SR} & 2 & 20,000 & UNK & 75\\

\hline
\hline

\citet{mulholland2013suicidal} & EN & SD  & Online DB & Song Lyrics & Manual & - & 2 & 810 & UNK & 37\\
\citet{zervopoulos2019language}  & EL & SDI & - & POEM & Manual & - & 2 & 90 & UNK & 12\\

\citet{reynolds2013depression} & CF & DP, AX \& PA & Tablets & - & - & - & - & - & UNK & 50 \\
\bottomrule
\end{tabular}
}
\caption{Overview of non-social media, text-based mental health datasets. The table summarizes each dataset’s language, mental disorder(s), platform, data form, annotation method, labeling instrument, availability, and citation count (till June 2025). 
\small \textit{Abbreviations:} SD - Suicidal, DP - Depression, MinDP - Minor Depression, PD - Postnatal Depression, PTSD - Post Traumatic Stress Disorder, OCD - Obsessive Compulsive Disorder, SMI - Severe Mental Illness, MDD - Major Depressive Disorder, MHD - Mental Health Distress, AjD - Adjustment Disorder, BD - Bipolar Disorder, SZ - Schizophrenia, AX - Anxiety, SDI - Suicidal Ideation, FEP - First Episode Psychosis, DEM - Dementia, PA - Panic Attack, PUB - Public, DUA- Data Use Agreement, RSTR - Restricted, UNK - Unknown, DS - Discharge Summaries}
\label{tab:mental_health_datasets}
\end{table*}

\begin{figure}[h]
\centering
\includegraphics[width=\columnwidth]{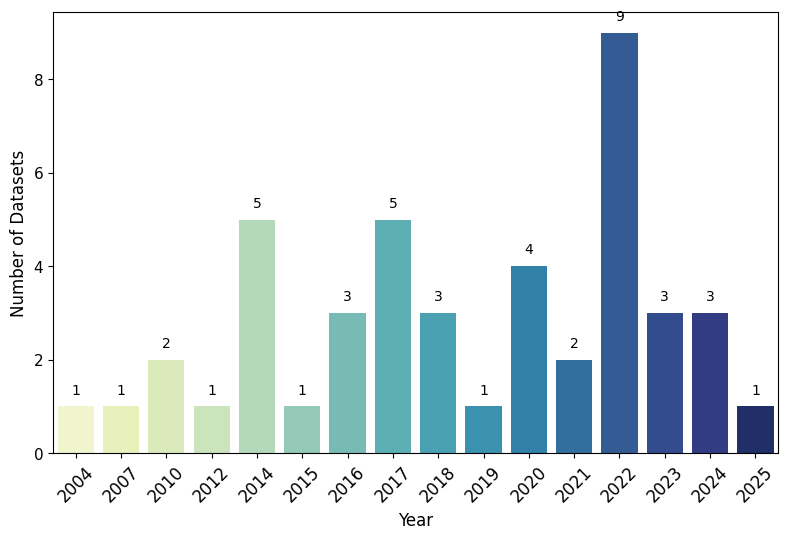}
\caption{Distribution of the datasets by year}
\label{fig: EDA by Year}
\end{figure}

\textbf{Temporal Distribution} Figure \ref{fig: EDA by Year} illustrates the number of free-text based datasets proposed and analyzed in published studies each year between 2004 and 2025. The introduction of such datasets remained minimal and steady up to 2012. In 2014 and 2017, there was a noticeable rise in the number of proposed datasets, indicating intensified research efforts in this domain during that period. The proposal of such datasets peaked in 2022, which can be attributed to greater attention to mental health concerns after the COVID-19 pandemic \cite{bucur2025state}. The lower count in 2025 is probably a result of the systematic search being conducted until June 2025. Despite this overall growth, the process of collecting such datasets often involves complex procedures, including ethics reviews, annotator training, and repeated permission requests, which may explain the limited number of datasets over the years.

\begin{figure}[h]
\centering
\includegraphics[width=.8\columnwidth]{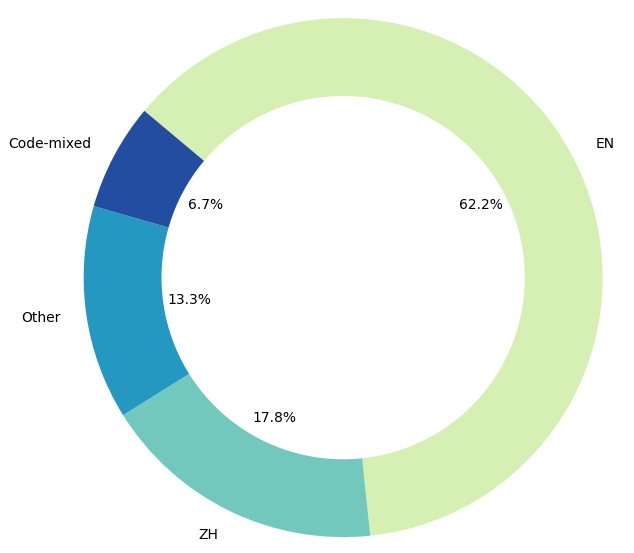}
\caption{Distribution of datasets into by language groups.}
\label{fig:EDA by Lang}
\end{figure}

\paragraph{Language Distribution} Figure \ref{fig:EDA by Lang} illustrates the distribution of the datasets by language group. English (EN) accounts for the majority, comprising 62.2\% of the datasets, highlighting its predominant role in this research area. Chinese (ZH) represents 17.8\% of the total. The `Other' group includes a variety of less-represented languages, such as Korean (KO), Polish (PL), Greek (EL), Japanese (JA), Thai (TH), and Arabic (AR); collectively representing another 13.3\%. Additionally, 6.4\% of the datasets fall under the “Code-mixed” category, which includes multilingual combinations like EN-ES (English-Spanish), EN-ZH (English-Chinese), and ES-CL (Spanish-Chilean). While there is some linguistic variety, English datasets dominate in mental health research, indicating a need for more inclusive and multilingual dataset development.

\paragraph{Mental Health Disorder Distribution} 



\begin{figure}[h]
\centering
\includegraphics[width=\columnwidth]{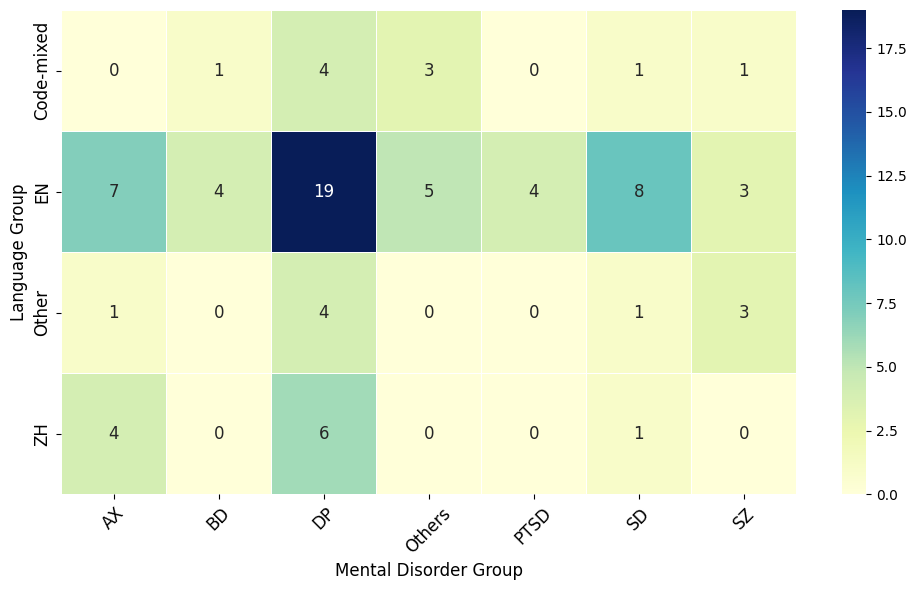}
\caption{Heatmap showing the distribution of datasets across language groups and disorders.}
\label{fig:heatmap}
\end{figure}

We consider the distribution of datasets for different mental health disorders and language diversity. In Figure \ref{fig:heatmap}, we present a heatmap with the distribution of mental health datasets across different disorders and languages. Since some datasets cover more than one disorder, the total count of disorders is higher than the number of datasets. To make things clearer, we group similar disorders; depression related conditions, grouped under `DP' (including Depression (DP), Major Depressive Disorder (MDD), Minor Depression (MinDP), and Postnatal Depression (PD)) are the most studied, appearing in 33 datasets. Anxiety (AX) and suicidal ideation/suicide(SD) follow with 12 and 11 datasets, respectively. Other less common disorders, such as Adjustment Disorder (AjD), Obsessive–compulsive disorder (OCD), Dysthymic Disorder (DD), and Dementia (DEM) are grouped under `Others'; each accounts for 8 datasets. Schizophrenia (SZ) and First-episode Psychosis (FEP), grouped together as `SZ', are analyzed in 5 datasets, while Post-Traumatic Stress Disorder (PTSD), and Bipolar Disorder (BD) are the least represented ones.

English datasets dominate the landscape, especially for depression (DP) and suicidal ideation (SD). Code-mixed and non-English data remain underrepresented across most disorders. This highlights a concentration on depression and a linguistic imbalance, emphasizing the need for more diverse mental health datasets.

\begin{figure}[!ht]
\centering
\includegraphics[width=\columnwidth]{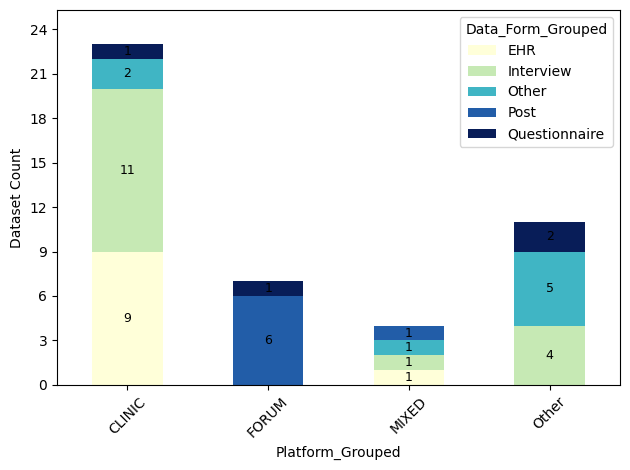}
\caption{Distribution of data types across platforms.}
\label{fig:platform vs dataform}
\end{figure}

\paragraph{Platform \& Data Type Distribution} Figure \ref{fig:platform vs dataform} shows the distribution of data types across four platform types: CLINIC, FORUM, MIXED (from multiple sources), and OTHER (including apps, virtual agents, online therapy chat, telemedicine platform and suicide notes). For simplicity, we group similar data types, like Electronic Health Records (EHR),
Electronic Medical Records (EMR), clinical records (CR), and Discharge Summaries (DS) are grouped under `EHR'; essays and questionnaires fall under `questionnaires' category; while types like interviews and posts are kept distinct.  

\noindent As shown in Figure \ref{fig:platform vs dataform}, clinical datasets mainly stem from interviews (48\%) and EHRs (39\%), reflecting structured clinical data sources. Forum datasets are heavily composed of user-generated posts (86\%). Mixed platform datasets are more balanced, with one dataset each from interviews, posts, questionnaires, and other forms (each 25\%). In the “Other” category, data forms are diverse: interviews and unconventional sources (like phone recordings, SMS, app data) both account for above 80\%, and questionnaires make up the remaining. 

\begin{figure}[h]
\centering
\includegraphics[width=\columnwidth]{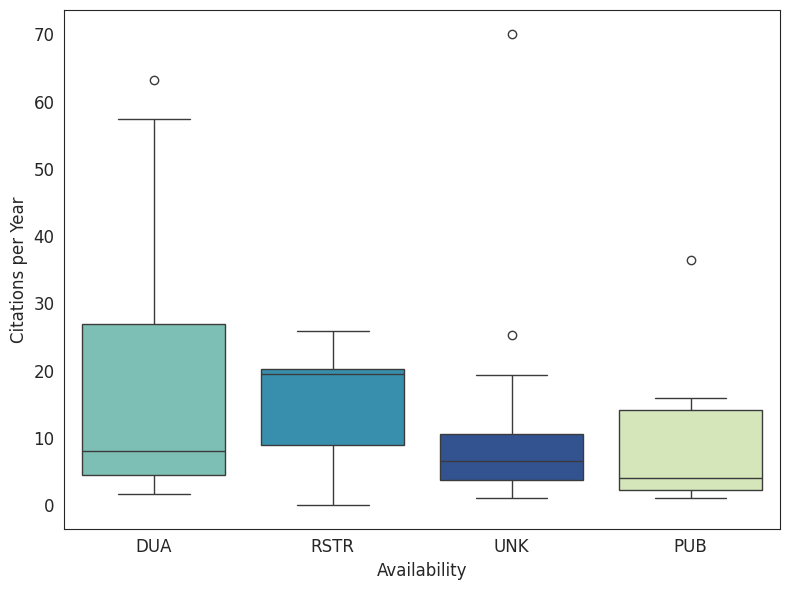}
\caption{Distribution of citation counts by dataset availability type.}
\label{fig:avail}
\end{figure}

\paragraph{Availability \& Impact}
Figure \ref{fig:avail} displays the distribution of citation counts (as per Google Scholar\footnote{\scriptsize\href{https://scholar.google.com/}{https://scholar.google.com/}}) for mental health datasets categorized by their availability: datasets that are publicly available (PUB), datasets that require a Data Use Agreement (DUA), datasets labeled as Restricted (RSTR), and datasets with unspecified availability (UNK). Citation counts are normalized by year to account for differences in publication age. Datasets categorized as DUA and RSTR generally have higher citation counts compared to those that are publicly accessible or have unclear availability. Notably, DUA datasets show a wider range and a higher median citation rate, while RSTR datasets demonstrate a more consistent citation pattern. In contrast, PUB and UNK datasets have lower medians and are more tightly clustered around fewer citations, though some outliers exist. 

\begin{figure}[h]
\centering
\includegraphics[width=.9\columnwidth]{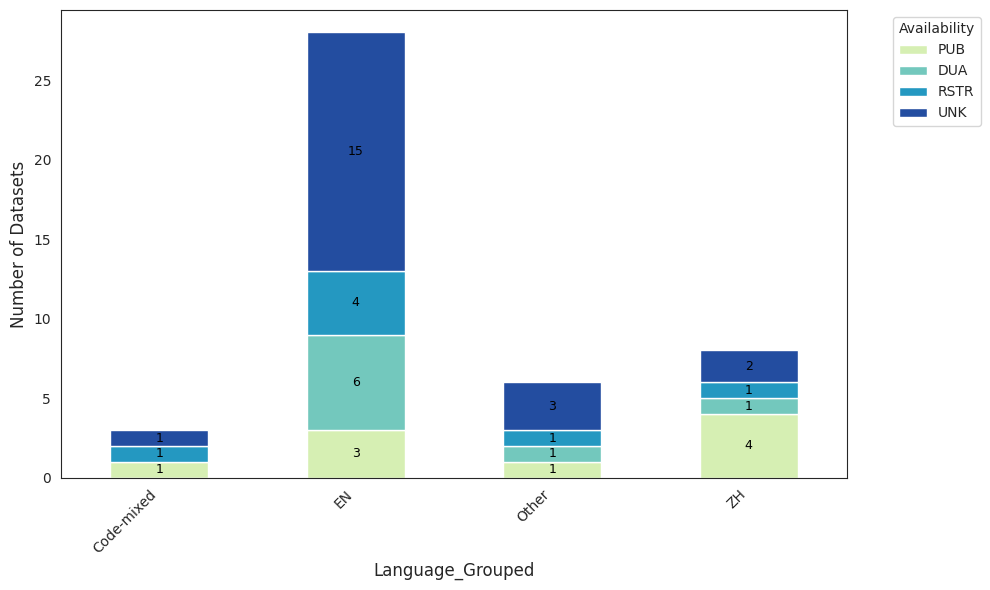}
\caption{Availability of datasets across language groups.}
\label{fig: availability vs lang}
\end{figure}

\noindent The citation trends can also be partly explained by language distribution (Figure \ref{fig: availability vs lang}). 

\noindent Many English datasets are in the UNK category, while Chinese datasets are more publicly available (PUB). However, English-language PUB datasets rarely include depression-related data, and none originate from clinical settings. In contrast, English-language DUA datasets predominantly focus on depression and are collected in clinical contexts -- this contributes to their higher credibility and research utility. Then again, most academic research is conducted in English; the higher citation rates of DUA and RSTR datasets reflect language dominance. Meanwhile, the lower citation rates for PUB datasets may be influenced by their association with less widely used languages like Chinese.

\begin{figure}[h]
\centering
\includegraphics[width=\columnwidth]{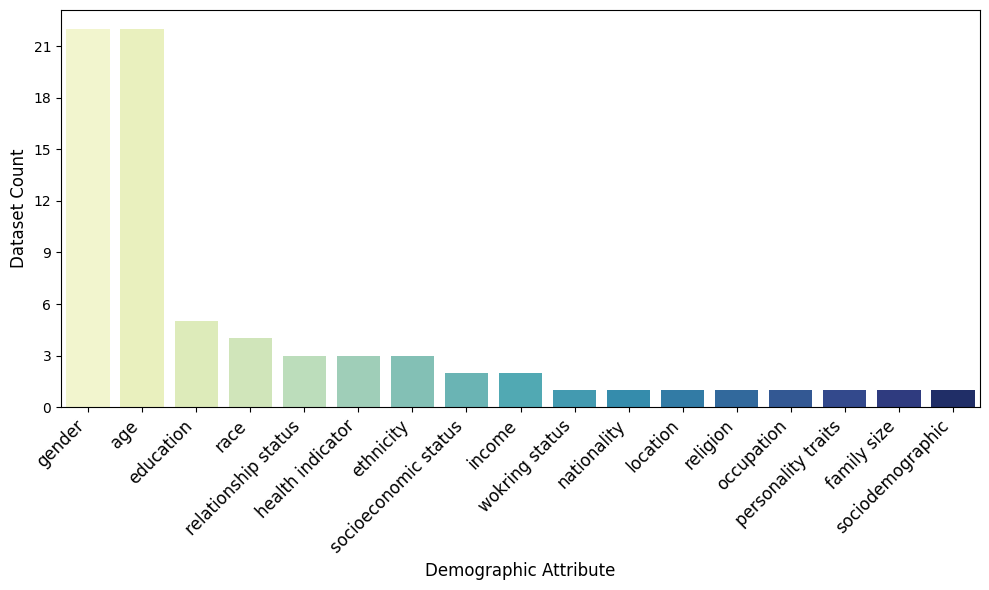}
\caption{Frequency of demographic attributes across datasets.}
\label{fig:EDA by demo}
\end{figure}

\paragraph{Demographic Representation} Figure \ref{fig:EDA by demo} shows how frequently different demographic attributes appear in the datasets shown in this survey. 

\noindent Demographic attributes like gender and age dominate, appearing in over 24 datasets, while others like education, race, and ethnicity are moderately included. Fewer datasets report attributes like income, relationship status, or health indicators (e.g., height, weight, blood pressure). Some authors \cite{hiraga2017predicting, low2010detection} use demographic information to select negative samples that match the same profile type as the positive ones, which is rarely done in social media-based data, resulting in poor performance on the minority class \cite{demo1, cao2024machine}. Some datasets focus on specific populations such as students \cite{shen2022automatic, rude2004language, nobles2018identification}, veterans \cite{gratch-distress, poulin2014predicting}, seafarers \cite{mo2024multimodal}, new mothers \cite{krishnamurti2022identification}, or politicians \cite{rheault2016expressions}. Some have used demographic features in classification tasks, often reporting improved model performance \cite{shin-dialogue, hcet2021, tasnim-depac}. 

\section{Tools \& Techniques for Data Annotation \& Labeling}

Most datasets in our review are manually annotated by annotators or expert clinicians, following widely used tools and techniques. This section describes these datasets and the tools employed. Some datasets, such as those by \citet{tyshchenko2018depression}, \citet{communityVScontrol}, \citet{hiraga2017predicting}, and \citet{alghamdi2020predicting}, are based on self-reported user content, typically collected from forums where individuals voluntarily share their thoughts under relevant discussion threads. 


\subsection{Clinical Diagnostic Instruments} 

Clinical diagnostic tools are assessments carried out by qualified healthcare professionals using standardized frameworks. They typically involve face-to-face evaluations, structured interviews, and expert judgment to ensure reliable and consistent mental health diagnoses. This systematic review highlights multiple studies that utilized clinical diagnostic tools for labeling and assessment.

The DSM-IV (Diagnostic and Statistical Manual of Mental Disorders, Fourth Edition) \cite{dsmiv} and its updated version, the DSM-V \cite{dsm5}, offer standardized criteria for diagnosing a wide range of mental health conditions. Diagnoses based on DSM are often recorded during clinical visits and stored in EHRs, making them a key source of labeled data. Several works in this survey rely on DSM-IV for annotation, including \citet{geraci2017applying}, \citet{he2017automated}, \citet{marrie2018effects} and \citet{wu2020using}, while studies such as \citet{xu2025identifying}, \citet{alghamdi2020predicting}, and \citet{oh2024development} utilize the updated DSM-V guidelines. Notably, \citet{aich2022towards} incorporated both frameworks to account for changes in diagnostic criteria across versions.

The ICD (International Classification of Diseases), particularly ICD-9 and ICD-10 \cite{world1992icd}, is used for coding clinical diagnoses, including mental and behavioral disorders. While the DSM focuses on mental health, the ICD covers all diseases. For instance, the code F32.9 in ICD-10 represents an unspecified single episode of major depressive disorder. In this survey, ICD-9 is used by \citet{hcet2021}, while \citet{xu2025identifying} and \citet{wawer2022single} utilize ICD-10 for data annotation, and \citet{ive2024data} uses both editions to label anxiety.

The LIFE (Longitudinal Interval Follow-Up Evaluation) is a structured method used to monitor the long-term progression of psychiatric conditions, typically every six months, particularly within longitudinal studies \cite{keller1987longitudinal}. \citet{brainsci12121717} underlines that LIFE is used in both clinical practice and research to assess the long-term impact of psychiatric disorders. In this survey, only \citet{low2010detection} uses LIFE to classify the interviews.

The CAPS (Clinician-Administered PTSD Scale) is a structured interview for diagnosing PTSD. It assesses 20 core PTSD symptoms, onset, duration, impairment, and dissociative features related to a specific traumatic event and is considered the gold standard for PTSD evaluation \cite{blake1995development}. Clinicians in \citet{he2017automated} use both the DSM-IV PTSD module and the CAPS scale to annotate the dataset with binary labels for PTSD presence or absence.

The HAMD (Hamilton Depression Rating Scale) \cite{RENEMANE2021175} is a clinician-administered scale used to assess depression severity through patient interviews, including 17 core items that rate depression from mild to severe. Similarly, HAMA (Hamilton Anxiety Rating Scale)  \cite{matza2010identifying} evaluates anxiety symptoms. In this survey, \citet{cmdc}, \citet{li2023detection}, \citet{shin2022detection} use HAMD to annotate the dataset for MDD or suicidal ideation, while \citet{jiang2022mmda} use both HAMD and HAMA for depression and anxiety labeling.

The MADRS (Montgomery-Åsberg Depression Rating Scale) \cite{MULLER2003255} is a clinician-administered tool for assessing depression severity, emphasizing core symptoms. It is often used in clinical trials and is sensitive to changes in symptoms. \citet{kangning} uses MADRS to identify depression from transcribed interviews.

\subsection{Screening Questionnaires and Self-Report Scales}

Screening questionnaires and self-report scales are tools where individuals assess their own mental health by answering standardized questions. These are often based on validated clinical criteria for quickly screening for symptoms. Several studies included in this review employ screening questionnaires and self-report scales for mental health disorder identification.

Some studies in this review, \citet{tasnim-depac}, \citet{cmdc}, \citet{alghamdi2020predicting}, \citet{hull2020two} and \citet{shin2022detection} use the Patient Health Questionnaire (PHQ-9) \cite{kroenke2001phq} for dataset annotations and defining gold standard labels for depression. \citet{ringeval2019avec} uses the older version of PHQ-9, PHQ-8, which consists of the same questions, excluding the suicidal ideation item \cite{kroenke2009phq}. \citet{rude2004language} employs the Inventory to Diagnose Depression–Lifetime (IDD-L) \cite{zimmerman1987inventory} to analyze college students' essays, identifying depression through their reflections on past experiences. The Beck Depression Inventory (BDI-II) \cite{beck1996beck}, a self-reported tool used to assess the intensity of depressive symptoms, \citet{rude2004language} employs both BDI-II and IDD-L for dataset annotation. In addition, the Beck Anxiety Inventory (BAI) \cite{beck1993beck} and the Beck Scale for Suicide Ideation (BSS) \cite{beck1991manual} are used to measure anxiety and suicidal thoughts, respectively, with \citet{shin2022detection} applying both to label transcribed interviews. In a study by \citet{krishnamurti2022identification}, the Edinburgh Postnatal Depression Scale (EPDS) \cite{cox1987detection} is used to identify depression in essays collected from pregnant women through an app. \citet{wu2020using} combines Sheehan Disability Scale \cite{sheehan1996measurement} with DSM-IV guidelines to identify major depressive disorder (MDD), schizophrenia (SZ), bipolar disorder (BD), and other psychiatric conditions from discharge summaries, extracted from patients’ EHRs; whereas \citet{shen2022automatic} uses Zung Self-Rating Depression Scale (SDS) \cite{TENTI2022158} to label depression from transcribed interviews collected through an app. \citet{tasnim-depac}, \citet{hull2020two} and \citet{mo2024multimodal} use the GAD-7 (Generalized Anxiety Disorder-7) \cite{10.1001/archinte.166.10.1092} to classify anxiety from their respective datasets. \citet{alghamdi2020predicting} uses the Quick Inventory of Depressive Symptomatology- Self-Report (QIDS-SR) \cite{RUSH2003573} tool to label depression in posts from a psychological forum in Arabic.

\subsection{Manual or Contextual Labeling}
Several datasets in this review do not explicitly mention the annotation tools used; however, they do indicate that trained annotators or clinicians conducted the labeling \cite{uzuner2017natural}. In some instances, studies report inter-rater agreement measures like Cohen’s Kappa (e.g. \citet{milne2016clpsych}, \citet{ghosh2020cease}, \citet{caicedo2020assessment}), Fleiss’s Kappa \cite{milne2016clpsych}, or Krippendorff’s~$\alpha$ \cite{pestian2012sentiment} to justify the labeling decisions, and occasionally a third annotator was used to resolve conflicts. A few datasets generate gold-standard labels using NLP-based methods such as Text Hunter \cite{jackson2017natural}, custom ontologies \cite{pestian2010suicide}, or frameworks like MALLET and LIWC \cite{howes2014linguistic}. \citet{hamalainen2021detecting} uses keyword search (i.e., ``depression", ``anxiety") to label the datasets. 

Few miscellaneous datasets contain song lyrics \cite{mulholland2013suicidal}, poems \cite{zervopoulos2019language} from artists who have committed suicide, even ancient texts scraped from Babylonian tablets \cite{reynolds2013depression}; the authors do not mention using any labeling techniques.

\subsection{Clinical Implications} 
Different labeling strategies in mental health datasets carry distinct clinical implications. Clinical diagnostic tools are widely considered the most reliable or gold standard \cite{arrow2023evaluating}, offering standardized and consistent evaluations that support accuracy across clinicians\footnote{\url{https://www.verywellmind.com/what-is-reliability-2795786}}. However, these tools have limitations, including time constraints (e.g., typical sessions lasting 15–20 minutes), limited accessibility (shortage of clinicians), and missed opportunities for monitoring between appointments (often referred to as ``clinical whitespace") \cite{stene2019contact}.

To address these gaps, self-report tools have become increasingly valuable. They capture patients’ first-hand perspectives and are ideal for digital use \cite{arrow2023evaluating}, especially when clinicians are unavailable, such as during clinical whitespace periods \cite{coppersmith2017scalable} or while patients await intake appointments \cite{cruz2013appointment}. While self-reports offer efficiency and scalability, they should not replace in-person evaluations, as they are susceptible to social desirability effects, recall bias \cite{althubaiti2016information}, and trust issues. Instead, they should serve as supplementary tools \cite{arrow2023evaluating}. In fact, research supports combining clinician ratings and self-reports for a more comprehensive understanding of patient conditions \cite{uher2012self}.

When clinical labels are unavailable, manual annotation, together with annotator agreement techniques, is often used to scale data labeling. While useful, these methods can suffer from reduced accuracy if not guided by trained professionals \cite{sylolypavan2023impact}. Overall, each labeling method presents trade-offs between scalability, clinical rigor, and data quality.

\section{Computational Modeling}
\label{sec:appendixA}

The studies reviewed in this systematic review have been used to develop mental health classification and prediction systems using a broad spectrum of approaches \cite{harrigian2021state,rissola2020beyond,bucur2025state}. While a thorough review of computational approaches is outside the scope of this survey, in this section, we provide the reader with a brief summary of the computational models applied to the task. Exploring computational models allows us to better understand prevailing methodological trends, evaluate benchmarks, and identify potential limitations or biases inherent in different approaches. This insight is vital for ensuring replicability and shaping future research directions.

Early works use manual or qualitative techniques while later studies applied traditional machine learning (ML) models like support vector machine (SVM) \cite{jackson2017natural, pestian2010suicide, pestian2012sentiment, he2017automated, uzuner2017natural, tyshchenko2018depression, nobles2018identification, diederich2007ex}, Logistic Regression \cite{pestian2012sentiment, hiraga2017predicting, li2023detection, aich2022towards, hcet2021, howes2014linguistic, mo2024multimodal, tasnim-depac}, Decision Trees \cite{he2017automated, zhou2015identifying, shen2022automatic}, Adaboost \cite{pestian2010suicide, uzuner2017natural, jiang2022mmda, mo2024multimodal}, XGboost \cite{wawer2022single, oh2024development, kangning}, and Naive Bayes \cite{schoene2016automatic, ghosh2020cease, he2017automated, hiraga2017predicting}. Traditional ML classifiers are often supported by feature engineering techniques such as Linguistic Inquiry and Word Count (LIWC) \cite{boyd2022development} analysis \cite{rude2004language, li2023detection, howes2014linguistic, nobles2018identification}, Term Frequency-Inverse Document Frequency (tf-idf) \cite{ive2024data, geraci2017applying, nobles2018identification}, or keyword extraction. Deep learning models, particularly Convolutional Neural Networks (CNN) \cite{ghosh2020cease, uzuner2017natural, tyshchenko2018depression}, Long Short-Term Memory (LSTM) \cite{hamalainen2021detecting, uzuner2017natural, kangning, krishnamurti2022identification}, and Gated Recurrent Unit (GRU) \cite{jackson2017natural, ghosh-etal-2021-detecting} are also used to capture nuanced patterns in unstructured text.

Some datasets have been used to develop interpretable models or hybrid systems that combine rule-based methods with ML (e.g., Conditional Random Field (CRF) \cite{pestian2012sentiment, wu2020using}, TextHunter + ConText \cite{jackson2017natural}). More recent work adopts transformer-based models like Clinical-BigBird \cite{ive2024data}, MentalBERT \cite{aich2022towards}, MentalRoBERTa \cite{aich2022towards}, and large language models (LLMs) like Qwen2-72B \cite{xu2025identifying}, highlighting a shift from fine-tuning pre-trained language models to using instruction-tuned LLMs for domain-specific tasks with little or no fine-tuning.

\section{Identified Trends \& Research Gaps}

This section summarizes the main findings of this review along with suggestions for future directions. Unsurprisingly, most available datasets are in English. Chinese follows in prevalence, while there are very few datasets in other languages, such as Korean, Arabic, or Polish. There are no datasets in low-resource languages. This gap may be partly due to the complexity of creating these datasets, particularly in clinical contexts. Collecting data often requires patient consent, approval from ethics boards, and substantial manual effort to annotate datasets.

Another important finding is that depression is the most studied mental health disorder within these datasets. Studies either focus exclusively on depression or include related conditions such as anxiety and bipolar disorder. Other conditions, such as PTSD, schizophrenia, and eating disorders, are not well represented. This highlights the need to develop datasets that encompass a broader range of mental health issues. On the other hand, compared to the number of social media datasets \cite{garg2023mental, bucur2025survey}, clinical datasets are scarce due to their sensitive nature. Releasing more public or even agreement-governed clinical datasets could expand research opportunities and improve reproducibility.



Moreover, the annotation procedures and labeling techniques vary across the datasets. Some use clinical diagnostic tools such as DSM and ICD, while others rely on self-reported questionnaires like PHQ and BDI. However, most papers do not explain why a specific tool was chosen, indicating a lack of reporting standards \cite{chancellor2020methods}. Additionally, some datasets are labeled manually or through keyword searches, and a few do not clarify their labeling methods at all. This lack of transparency can make it difficult to trust or replicate the research findings.

Advancements in NLP offer promising solutions to the challenges posed by inconsistent and opaque labeling practices in mental health datasets. With instruction-tuned language models and structured prompting techniques, NLP can support (semi-)automation and standardization of labeling based on formal diagnostic criteria such as DSM or ICD. Approaches like chain-of-thought (CoT) prompting \cite{wei2022chain} can emulate clinical reasoning and even improve labeling accuracy \cite{shi2024enhancing}, while chain-of-empathy (CoE) \cite{lee2023chain} frameworks are well-suited for understanding emotionally nuanced texts, such as therapy transcripts or suicide notes. These strategies can enhance both the consistency and interpretability of labels, enabling the creation of scalable and clinically relevant datasets, even when direct clinician involvement is limited.

\section{Conclusion \& Future Directions}

In this paper, we presented the first systematic review of mental health free-text datasets beyond social media. We analyze these datasets with respect to different dimensions such as their distribution in terms of languages and mental health disorders, the data types included in them, and their availability. 

We revisit the research questions (RQ) posed in the introduction (Section \ref{sec:intro}) and present the main findings of our review below:

\paragraph{RQ\textsubscript{1}:}What non-social media free-text based datasets are available for mental health research, and how do they vary by source, structure, and population? 

\begin{resultbox}
\textbf{RQ\textsubscript{1} Findings:} \textit{Most datasets are in English and primarily focus on depression, while other languages and mental health conditions remain underrepresented. More Chinese datasets are publicly available than English ones. In terms of data collection, some actively balance samples by selecting equal numbers from the control group matching the same profile type. The nature of text-based data also varies widely, ranging from phone recordings and messages to essays, interview transcripts, and suicide notes, mostly collected in clinical settings.}
\end{resultbox}

\paragraph{RQ\textsubscript{2}:}How are mental health conditions defined and labeled in these datasets, and what are the clinical implications of these labeling methods?

\begin{resultbox}
\textbf{RQ\textsubscript{2} Findings:}\textit{ Clinical diagnostic tools, screening questionnaires, self-report scales, and manual contextual labeling have been used to annotate mental health disorders. These labeling techniques can complement one another and collectively serve as gold standards.}
\end{resultbox}

\paragraph{RQ\textsubscript{3}:}What factors contribute to the popularity and adoption of non-social media free-text based mental health datasets in research?

\begin{resultbox}
\textbf{RQ\textsubscript{3} Findings:}\textit{ Along with dataset availability, factors such as the data collection setting (clinical vs. informal), language, data type (interviews, EHRs, essays), mental health disorders covered, and the credibility and reliability of the labeling and annotation methods all contribute to the popularity and adoption of non-social media free-text based mental health datasets in research.}
\end{resultbox}


\noindent There is significant room for improvement in the development and use of these types of datasets. Expanding dataset creation to include more languages and geographic regions can help address current imbalances. Additionally, increasing the representation of a broader range of mental health disorders, not just depression, would make research findings more comprehensive. Establishing clear and consistent labeling methods is essential for ensuring reproducibility and building trust in research outcomes, and advances in NLP, particularly in explainable AI, can help bridge these gaps. Lastly, improving access to high-quality datasets, while maintaining ethical and privacy standards, can facilitate collaborative research.

\section*{Limitations}

In this systematic review, we used the PRISMA methodology and conducted a comprehensive literature search using the Publish or Perish software. This approach ensured thorough coverage of datasets on mental health disorders across both the NLP and clinical domains. Although we used mental health-related terms for our keyword searches, it is possible that studies may use alternative terms or less common phrases to describe mental health disorders, which may not have been included in our search strategy. As a result, some relevant works might have been unintentionally overlooked. Additionally, the search queries were in English, which may have excluded relevant non-English publications.  Finally, Publish or Perish does not index certain databases, such as CINAHL, potentially limiting coverage of some clinically oriented studies.

\section*{Ethics Statement}

This systematic review draws from previously published studies to guide future research in identifying mental health disorders beyond social media. While we present the available datasets used for this purpose in our review, we did not make any attempts to build prediction models using this data. We acknowledge that detecting early signs of mental health disorders requires adherence to ethical protocols. 
Moreover, misuse of these sensitive data or of the models trained on the data can lead to stigmatization or harm to individuals with mental health disorders \cite{chancellor2019taxonomy}. 

\section*{Acknowledgements}

We would like to thank the anonymous reviewers for their constructive feedback and thoughtful suggestions, which helped improve the clarity and quality of this work. We are also grateful to our collaborators for their valuable contributions and insightful discussions throughout the project. Finally, we acknowledge the creators and maintainers of the datasets studied in our review for developing and curating these resources, thus enabling and advancing research in this domain.


\section*{References}
\label{sec:reference}
\vspace{-7mm}
\bibliography{anthology,custom,languageresource}

@article{calvo2017natural,
  title={Natural language processing in mental health applications using non-clinical texts},
  author={Calvo, Rafael A and Milne, David N and Hussain, M Sazzad and Christensen, Helen},
  journal={Natural Language Engineering},
  volume={23},
  number={5},
  pages={649--685},
  year={2017},
  publisher={Cambridge University Press}
}

@article{islam2018depression,
  title={Depression detection from social network data using machine learning techniques},
  author={Islam, Md Rafiqul and Kabir, Muhammad Ashad and Ahmed, Ashir and Kamal, Abu Raihan M and Wang, Hua and Ulhaq, Anwaar},
  journal={Health information science and systems},
  volume={6},
  pages={1--12},
  year={2018},
  publisher={Springer}
}

@article{boettcher2021studies,
  title={Studies of depression and anxiety using reddit as a data source: scoping review},
  author={Boettcher, Nick},
  journal={JMIR mental health},
  volume={8},
  number={11},
  pages={e29487},
  year={2021}
}

@article{adil2022corpus,
  title={A corpus-based stylistic analysis of online suicide notes retrieved from Reddit},
  author={Adil Jaafar, Eman and Abdul-Salam Jasim, Haya},
  journal={Cogent Arts \& Humanities},
  volume={9},
  number={1},
  pages={2047434},
  year={2022},
  publisher={Taylor \& Francis}
}

@inproceedings{chatterjee2022suicide,
  title={Suicide ideation detection using multiple feature analysis from Twitter data},
  author={Chatterjee, Moumita and Samanta, Poulomi and Kumar, Piyush and Sarkar, Dhrubasish},
  booktitle={IEEE DELCON},
  year={2022}
}

@article{khafaga2023deep,
  title={Deep learning for depression detection using Twitter data},
  author={Khafaga, D Sami and Auvdaiappan, Maheshwari and Deepa, K and Abouhawwash, Mohamed and Karim, F Khalid},
  journal={Intelligent Automation \& Soft Computing},
  volume={36},
  number={2},
  pages={1301--1313},
  year={2023}
}

@article{bucur2025state,
  author={Bucur, Ana-Maria and Moldovan, Andreea-Codrina and Parvatikar, Krutika and Zampieri, Marcos and KhudaBukhsh, Ashiqur R. and Dinu, Liviu P.},
  journal={IEEE Journal of Biomedical and Health Informatics}, 
  title={On the State of NLP Approaches to Modeling Depression in Social Media: A Post-COVID-19 Outlook}, 
  year={2025},
  volume={29},
  number={6},
  pages={4439-4451}
  }

@inproceedings{raihan2024mentalhelp,
  title={Mentalhelp: A multi-task dataset for mental health in social media},
  author={Raihan, Nishat and Puspo, Sadiya Sayara Chowdhury and Farabi, Shafkat and Bucur, Ana-Maria and Ranasinghe, Tharindu and Zampieri, Marcos},
  booktitle={Proceedings of LREC-COLING},
  year={2024}
}

@inproceedings{bucur2025datasets,
  title={Datasets for depression modeling in social media: An overview},
  author={Bucur, Ana-Maria and Moldovan, Andreea and Parvatikar, Krutika and Zampieri, Marcos and Khudabukhsh, Ashiqur and Dinu, Liviu P},
  booktitle={Proceedings of CLPsych},
  year={2025}
}

@article{boyd2022development,
  title={The development and psychometric properties of LIWC-22},
  author={Boyd, Ryan L and Ashokkumar, Ashwini and Seraj, Sarah and Pennebaker, James W},
  journal={Austin, TX: University of Texas at Austin},
  volume={10},
  number={1-47},
  pages={6},
  year={2022}
}

@inproceedings{raihan2026large,
  title={Large Language Models for Mental Health: A Multilingual Evaluation},
  author={Raihan, Nishat and Puspo, Sadiya Sayara Chowdhury and Bucur, Ana-Maria and Chancellor, Stevie and Zampieri, Marcos},
  booktitle={Proceedings of LowResLM},
  year={2026}
}

@article{abdulsalam2024suicidal,
  title={Suicidal ideation detection on social media: A review of machine learning methods},
  author={Abdulsalam, Asma and Alhothali, Areej},
  journal={Social Network Analysis and Mining},
  volume={14},
  number={1},
  pages={188},
  year={2024},
  publisher={Springer}
}

@article{skaik2020using,
  title={Using social media for mental health surveillance: a review},
  author={Skaik, Ruba and Inkpen, Diana},
  journal={ACM Computing Surveys},
  volume={53},
  number={6},
  pages={1--31},
  year={2020}
}

@inproceedings{rissola2020beyond,
	title        = {Beyond modelling: understanding mental disorders in online social media},
	author       = {Ríssola, Esteban Andrés and Aliannejadi, Mohammad and Crestani, Fabio},
	year         = 2020,
	booktitle    = {Proc. of ECIR},
	pages        = {296--310},
	organization = {Springer}
}

@inproceedings{harrigian2021state,
	title        = {On the State of Social Media Data for Mental Health Research},
	author       = {Harrigian, Keith and Aguirre, Carlos and Dredze, Mark},
	year         = 2021,
	booktitle    = {Proceedings of CLPsych}
}

@book{world1992icd,
  title={The ICD-10 classification of mental and behavioural disorders: clinical descriptions and diagnostic guidelines},
  author={{World Health Organization}},
  volume={1},
  year={1992},
  publisher={World Health Organization}
}

@article{kroenke2009phq,
  title={The PHQ-8 as a measure of current depression in the general population},
  author={Kroenke, Kurt and Strine, Tara W and Spitzer, Robert L and Williams, Janet BW and Berry, Joyce T and Mokdad, Ali H},
  journal={Journal of Affective Disorders},
  volume={114},
  number={1-3},
  pages={163--173},
  year={2009},
  publisher={Elsevier}
}

@article{beck1991manual,
  title={Manual for the Beck scale for suicide ideation},
  author={Beck, Aaron T and Steer, Robert A},
  journal={San Antonio, TX: Psychological Corporation},
  volume={63},
  year={1991}
}

@book{dsmiv,
  title={Diagnostic and Statistical Manual of Mental Disorders, Fourth Edition (DSM-IV)},
  author={APA},
  year={1994},
  publisher={American Psychiatric Association},
}

@book{dsm5,
  title={Diagnostic and Statistical Manual of Mental Disorders},
  edition={5th ed.},
  author={APA},
  publisher={American Psychiatric Publishing},
  year={2013},
}

@article{kroenke2001phq,
  title={The PHQ-9: validity of a brief depression severity measure},
  author={Kroenke, Kurt and Spitzer, Robert L and Williams, Janet BW},
  journal={Journal of general internal medicine},
  volume={16},
  number={9},
  pages={606--613},
  year={2001},
  publisher={Wiley Online Library}
}

@article{zimmerman1987inventory,
  title={The inventory to diagnose depression, lifetime version},
  author={Zimmerman, Mark and Coryell, William},
  journal={Acta Psychiatrica Scandinavica},
  volume={75},
  number={5},
  pages={495--499},
  year={1987},
  publisher={Wiley Online Library}
}

@article{beck1996beck,
  title={Beck depression inventory--II},
  author={Beck, Aaron T and Steer, Robert A and Brown, Gregory},
  journal={Psychological assessment},
  year={1996}
}

@article{keller1987longitudinal,
  title={The longitudinal interval follow-up evaluation: A comprehensive method for assessing outcome in prospective longitudinal studies},
  author={Keller, Martin B and Lavori, Philip W and Friedman, Barbara and Nielsen, Eileen and Endicott, Jean and McDonald-Scott, Pat and Andreasen, Nancy C},
  journal={Archives of general psychiatry},
  volume={44},
  number={6},
  pages={540--548},
  year={1987},
  publisher={American Medical Association}
}

@Article{brainsci12121717,
AUTHOR = {Porter, Richard J. and Moot, Will and Inder, Maree L. and Crowe, Marie T. and Douglas, Katie M. and Carter, Janet D. and Frampton, Christopher},
TITLE = {Validation of the Longitudinal Interval Follow-Up Evaluation for the Long-Term Measurement of Mood Symptoms in Bipolar Disorder},
JOURNAL = {Brain Sciences},
VOLUME = {12},
YEAR = {2022},
NUMBER = {12}
}

@article{cox1987detection,
  title={Detection of postnatal depression: development of the 10-item Edinburgh Postnatal Depression Scale},
  author={Cox, John L and Holden, Jeni M and Sagovsky, Ruth},
  journal={The British journal of psychiatry},
  volume={150},
  number={6},
  pages={782--786},
  year={1987},
  publisher={Cambridge University Press}
}

@article{blake1995development,
  title={The development of a clinician-administered PTSD scale},
  author={Blake, Dudley David and Weathers, Frank W and Nagy, Linda M and Kaloupek, Danny G and Gusman, Fred D and Charney, Dennis S and Keane, Terence M},
  journal={Journal of traumatic stress},
  volume={8},
  pages={75--90},
  year={1995},
  publisher={Springer}
}

@article{sheehan1996measurement,
  title={The measurement of disability},
  author={Sheehan, David V and Harnett-Sheehan, Kathy and Raj, BA8923116},
  journal={International clinical psychopharmacology},
  volume={11},
  pages={89--95},
  year={1996},
  publisher={LWW}
}

@article{10.1001/archinte.166.10.1092,
    author = {Spitzer, Robert L. and Kroenke, Kurt and Williams, Janet B. W. and Löwe, Bernd},
    title = {A Brief Measure for Assessing Generalized Anxiety Disorder: The GAD-7},
    journal = {Archives of Internal Medicine},
    volume = {166},
    number = {10},
    pages = {1092-1097},
    year = {2006},
}

@incollection{RENEMANE2021175,
title = {Chapter 17 - Hamilton depression rating scale: Uses and applications},
editor = {Colin R. Martin and Lan-Anh Hunter and Vinood B. Patel and Victor R. Preedy and Rajkumar Rajendram},
booktitle = {The Neuroscience of Depression},
publisher = {Academic Press},
pages = {175-183},
year = {2021},
author = {Lubova Renemane and Jelena Vrublevska}
}

@article{RUSH2003573,
title = {The 16-Item quick inventory of depressive symptomatology (QIDS), clinician rating (QIDS-C), and self-report (QIDS-SR): a psychometric evaluation in patients with chronic major depression},
journal = {Biological Psychiatry},
volume = {54},
number = {5},
pages = {573-583},
year = {2003},
author = {A.John Rush and Madhukar H Trivedi and Hicham M Ibrahim and Thomas J Carmody and Bruce Arnow and Daniel N Klein and John C Markowitz and Philip T Ninan and Susan Kornstein and Rachel Manber and Michael E Thase and James H Kocsis and Martin B Keller},
}

@article{beck1993beck,
  title={Beck anxiety inventory},
  author={Beck, Aaron T and Epstein, Norman and Brown, Gary and Steer, Robert},
  journal={Journal of consulting and clinical psychology},
  year={1993}
}

@article{TENTI2022158,
title = {A Narrative Review of the Assessment of Depression in Chronic Pain},
journal = {Pain Management Nursing},
volume = {23},
number = {2},
pages = {158-167},
year = {2022},
author = {Michael Tenti and William Raffaeli and Paola Gremigni}
}

@article{MULLER2003255,
title = {Differentiating moderate and severe depression using the Montgomery–Åsberg depression rating scale (MADRS)},
journal = {Journal of Affective Disorders},
volume = {77},
number = {3},
pages = {255-260},
year = {2003},
author = {Matthias J. Müller and Hubertus Himmerich and Barbara Kienzle and Armin Szegedi}
}

@inproceedings{ringeval2019avec,
  title={AVEC 2019 workshop and challenge: state-of-mind, detecting depression with AI, and cross-cultural affect recognition},
  author={Ringeval, Fabien and Schuller, Bj{\"o}rn and Valstar, Michel and Cummins, Nicholas and Cowie, Roddy and Tavabi, Leili and Schmitt, Maximilian and Alisamir, Sina and Amiriparian, Shahin and Messner, Eva-Maria and others},
  booktitle={Proceedings of AVEC},
  year={2019}
}

@article{jackson2017natural,
  title={Natural language processing to extract symptoms of severe mental illness from clinical text: the Clinical Record Interactive Search Comprehensive Data Extraction (CRIS-CODE) project},
  author={Jackson, Richard G and Patel, Rashmi and Jayatilleke, Nishamali and Kolliakou, Anna and Ball, Michael and Gorrell, Genevieve and Roberts, Angus and Dobson, Richard J and Stewart, Robert},
  journal={BMJ open},
  volume={7},
  number={1},
  pages={e012012},
  year={2017},
  publisher={British Medical Journal Publishing Group}
}

@article{pestian2010suicide,
  title={Suicide note classification using natural language processing: A content analysis},
  author={Pestian, John and Nasrallah, Henry and Matykiewicz, Pawel and Bennett, Aurora and Leenaars, Antoon},
  journal={Biomedical informatics insights},
  volume={3},
  pages={BII--S4706},
  year={2010},
  publisher={SAGE Publications Sage UK: London, England}
}

@article{pestian2012sentiment,
  title={Sentiment analysis of suicide notes: A shared task},
  author={Pestian, John P and Matykiewicz, Pawel and Linn-Gust, Michelle and South, Brett and Uzuner, Ozlem and Wiebe, Jan and Cohen, K Bretonnel and Hurdle, John and Brew, Christopher},
  journal={Biomedical informatics insights},
  volume={5},
  pages={BII--S9042},
  year={2012},
  publisher={SAGE Publications Sage UK: London, England}
}

@ARTICLE{communityVScontrol,
  author={Nguyen, Thin and Phung, Dinh and Dao, Bo and Venkatesh, Svetha and Berk, Michael},
  journal={IEEE Transactions on Affective Computing}, 
  title={Affective and Content Analysis of Online Depression Communities}, 
  year={2014},
  volume={5},
  number={3},
  pages={217-226},
  }

@inproceedings{milne2016clpsych,
  title={Clpsych 2016 shared task: Triaging content in online peer-support forums},
  author={Milne, David N and Pink, Glen and Hachey, Ben and Calvo, Rafael A},
  booktitle={Proceedings of the third workshop on computational linguistics and clinical psychology},
  pages={118--127},
  year={2016}
}

@article{he2017automated,
  title={Automated assessment of patients’ self-narratives for posttraumatic stress disorder screening using natural language processing and text mining},
  author={He, Qiwei and Veldkamp, Bernard P and Glas, Cees AW and de Vries, Theo},
  journal={Assessment},
  volume={24},
  number={2},
  pages={157--172},
  year={2017},
  publisher={Sage Publications Sage CA: Los Angeles, CA}
}

@article{wu2020using,
  title={Using text mining to extract depressive symptoms and to validate the diagnosis of major depressive disorder from electronic health records},
  author={Wu, Chi-Shin and Kuo, Chian-Jue and Su, Chu-Hsien and Wang, Shi-Heng and Dai, Hong-Jie},
  journal={Journal of affective disorders},
  volume={260},
  pages={617--623},
  year={2020},
  publisher={Elsevier}
}

@article{geraci2017applying,
  title={Applying deep neural networks to unstructured text notes in electronic medical records for phenotyping youth depression},
  author={Geraci, Joseph and Wilansky, Pamela and de Luca, Vincenzo and Roy, Anvesh and Kennedy, James L and Strauss, John},
  journal={BMJ Ment Health},
  volume={20},
  number={3},
  pages={83--87},
  year={2017},
  publisher={Royal College of Psychiatrists}
}

@article{tyshchenko2018depression,
  title={Depression and anxiety detection from blog posts data},
  author={Tyshchenko, Yevhen},
  journal={Nature Precis. Sci., Inst. Comput. Sci., Univ. Tartu, Tartu, Estonia},
  pages={6--46},
  year={2018}
}

@inproceedings{hiraga2017predicting,
  title={Predicting depression for japanese blog text},
  author={Hiraga, Misato},
  booktitle={Proceedings of ACL},
  year={2017}
}

@ARTICLE{kangning,
  author={Mao, Kaining and Wang, Deborah Baofeng and Zheng, Tiansheng and Jiao, Rongqi and Zhu, Yanhui and Wu, Bin and Qian, Lei and Lyu, Wei and Chen, Jie and Ye, Minjie},
  journal={IEEE Transactions on Biomedical Circuits and Systems}, 
  title={Analysis of Automated Clinical Depression Diagnosis in a Chinese Corpus}, 
  year={2023},
  volume={17},
  number={5},
  pages={1135-1152}
}

@article{rude2004language,
  title={Language use of depressed and depression-vulnerable college students},
  author={Rude, Stephanie and Gortner, Eva-Maria and Pennebaker, James},
  journal={Cognition \& Emotion},
  volume={18},
  number={8},
  pages={1121--1133},
  year={2004},
  publisher={Taylor \& Francis}
}

@article{reynolds2013depression,
  title={Depression and anxiety in Babylon},
  author={Reynolds, Edward H and Wilson, James V Kinnier},
  journal={Journal of the Royal Society of Medicine},
  volume={106},
  number={12},
  pages={478--481},
  year={2013},
  publisher={SAGE Publications Sage UK: London, England}
}

@article{li2023detection,
  title={Detection of suicidal ideation in clinical interviews for depression using natural language processing and machine learning: cross-sectional study},
  author={Li, Tim MH and Chen, Jie and Law, Framenia OC and Li, Chun-Tung and Chan, Ngan Yin and Chan, Joey WY and Chau, Steven WH and Liu, Yaping and Li, Shirley Xin and Zhang, Jihui and others},
  journal={JMIR medical informatics},
  volume={11},
  number={1},
  pages={e50221},
  year={2023},
  publisher={JMIR Publications Inc., Toronto, Canada}
}

@article{shin2022detection,
  title={Detection of depression and suicide risk based on text from clinical interviews using machine learning: possibility of a new objective diagnostic marker},
  author={Shin, Daun and Kim, Kyungdo and Lee, Seung-Bo and Lee, Changwoo and Bae, Ye Seul and Cho, Won Ik and Kim, Min Ji and Hyung Keun Park, C and Chie, Eui Kyu and Kim, Nam Soo and others},
  journal={Frontiers in psychiatry},
  volume={13},
  pages={801301},
  year={2022},
  publisher={Frontiers Media SA}
}

@article{low2010detection,
  title={Detection of clinical depression in adolescents’ speech during family interactions},
  author={Low, Lu-Shih Alex and Maddage, Namunu C and Lech, Margaret and Sheeber, Lisa B and Allen, Nicholas B},
  journal={IEEE transactions on biomedical engineering},
  volume={58},
  number={3},
  pages={574--586},
  year={2010},
  publisher={IEEE}
}

@article{krishnamurti2022identification,
  title={Identification of maternal depression risk from natural language collected in a mobile health app},
  author={Krishnamurti, Tamar and Allen, Kristen and Hayani, Laila and Rodriguez, Samantha and Davis, Alexander L},
  journal={Procedia computer science},
  volume={206},
  pages={132--140},
  year={2022},
  publisher={Elsevier}
}

@ARTICLE{cmdc,
  author={Zou, Bochao and Han, Jiali and Wang, Yingxue and Liu, Rui and Zhao, Shenghui and Feng, Lei and Lyu, Xiangwen and Ma, Huimin},
  journal={IEEE Transactions on Affective Computing}, 
  title={Semi-Structural Interview-Based Chinese Multimodal Depression Corpus Towards Automatic Preliminary Screening of Depressive Disorders}, 
  year={2023},
  volume={14},
  number={4},
  pages={2823-2838}
 }

@inproceedings{aich2022towards,
  title={Towards intelligent clinically-informed language analyses of people with bipolar disorder and schizophrenia},
  author={Aich, Ankit and Quynh, Avery and Badal, Varsha and Pinkham, Amy and Harvey, Philip and Depp, Colin and Parde, Natalie},
  booktitle={Findings of EMNLP},
  year={2022}
}

@ARTICLE{hcet2021,
  author={Meng, Yiwen and Speier, William and Ong, Michael and Arnold, Corey W.},
  journal={IEEE Journal of Biomedical and Health Informatics}, 
  title={HCET: Hierarchical Clinical Embedding With Topic Modeling on Electronic Health Records for Predicting Future Depression}, 
  year={2021},
  volume={25},
  number={4},
  pages={1265-1272}
  }

@inproceedings{jiang2022mmda,
  title={MMDA: A Multimodal Dataset for Depression and Anxiety Detection},
  author={Jiang, Yueqi and Zhang, Ziyang and Sun, Xiao},
  booktitle={Proceedings of ICPR},
  year={2022}
}

@inproceedings{howes2014linguistic,
  title={Linguistic indicators of severity and progress in online text-based therapy for depression},
  author={Howes, Christine and Purver, Matthew and McCabe, Rose},
  booktitle={Proceedings of CLPsych},
  year={2014}
}

@incollection{zhou2015identifying,
  title={Identifying patients with depression using free-text clinical documents},
  author={Zhou, Li and Baughman, Amy W and Lei, Victor J and Lai, Kenneth H and Navathe, Amol S and Chang, Frank and Sordo, Margarita and Topaz, Maxim and Zhong, Feiran and Murrali, Madhavan and others},
  booktitle={MEDINFO},
  year={2015}
}

@inproceedings{rheault2016expressions,
  title={Expressions of anxiety in political texts},
  author={Rheault, Ludovic},
  booktitle={Proceedings of NLP-CSS},
  year={2016}
}

@article{mo2024multimodal,
  title={A multimodal data-driven framework for anxiety screening},
  author={Mo, Haimiao and Hui, Siu Cheung and Liao, Xiao and Li, Yuchen and Zhang, Wei and Ding, Shuai},
  journal={IEEE Transactions on Instrumentation and Measurement},
  volume={73},
  pages={1--13},
  year={2024},
  publisher={IEEE}
}

@article{xu2025identifying,
  title={Identifying Psychiatric Manifestations in Outpatients with Depression and Anxiety: A Large Language Model-Based Approach},
  author={Xu, Shihao and Yan, Yiming and Ding, Yanli and Li, Feng and Zhang, Shu and Tang, Haoyun and Luo, Chao and Li, Yan and Liu, Hao and Mei, Yu and others},
  journal={medRxiv},
  pages={2025--01},
  year={2025},
  publisher={Cold Spring Harbor Laboratory Press}
}

@article{ive2024data,
  title={A Data-Centric Approach to Detecting and Mitigating Demographic Bias in Pediatric Mental Health Text: A Case Study in Anxiety Detection},
  author={Ive, Julia and Bondaronek, Paulina and Yadav, Vishal and Santel, Daniel and Glauser, Tracy and Cheng, Tina and Strawn, Jeffrey R and Agasthya, Greeshma and Tschida, Jordan and Choo, Sanghyun and others},
  journal={arXiv preprint arXiv:2501.00129},
  year={2024}
}

@inproceedings{shen2022automatic,
  title={Automatic depression detection: An emotional audio-textual corpus and a gru/bilstm-based model},
  author={Shen, Ying and Yang, Huiyu and Lin, Lin},
  booktitle={IEEE ICASSP},
  year={2022}
}

@article{wawer2022single,
  title={Single and cross-disorder detection for autism and schizophrenia},
  author={Wawer, Aleksander and Chojnicka, Izabela and Okruszek, Lukasz and Sarzynska-Wawer, Justyna},
  journal={Cognitive Computation},
  volume={14},
  number={1},
  pages={461--473},
  year={2022},
  publisher={Springer}
}

@inproceedings{zervopoulos2019language,
  title={Language processing for predicting suicidal tendencies: a case study in greek poetry},
  author={Zervopoulos, Alexandros Dimitrios and Geramanis, Evangelos and Toulakis, Alexandros and Papamichail, Asterios and Triantafylloy, Dimitrios and Tasoulas, Theofanis and Kermanidis, Katia},
  booktitle={IFIP AIAI},
  year={2019}
}

@inproceedings{tasnim-depac,
    title = "{DEPAC}: a Corpus for Depression and Anxiety Detection from Speech",
    author = "Tasnim, Mashrura  and
      Ehghaghi, Malikeh  and
      Diep, Brian  and
      Novikova, Jekaterina",
    booktitle = "Proceedings of CLPsych",
    year = "2022",

}

@inproceedings{schoene2016automatic,
  title={Automatic identification of suicide notes from linguistic and sentiment features},
  author={Schoene, Annika Marie and Dethlefs, Nina},
  booktitle={Proceedings of SIGHUM, LaTeCH},
  year={2016}
}

@inproceedings{mulholland2013suicidal,
  title={Suicidal tendencies: The automatic classification of suicidal and non-suicidal lyricists using nlp},
  author={Mulholland, Matthew and Quinn, Joanne},
  booktitle={Proceedings of IJCNLP},
  year={2013}
}

@inproceedings{ghosh2020cease,
  title={Cease, a corpus of emotion annotated suicide notes in English},
  author={Ghosh, Soumitra and Ekbal, Asif and Bhattacharyya, Pushpak},
  booktitle={Proceedings of LREC},
  year={2020}
}

@article{caicedo2020assessment,
  title={Assessment of supervised classifiers for the task of detecting messages with suicidal ideation},
  author={Caicedo, Roberto Wellington Acu{\~n}a and Soriano, Jos{\'e} Manuel G{\'o}mez and Sasieta, H{\'e}ctor Andr{\'e}s Melgar},
  journal={Heliyon},
  volume={6},
  number={8},
  year={2020},
  publisher={Elsevier}
}

@inproceedings{hamalainen2021detecting,
    title = "Detecting Depression in {T}hai Blog Posts: a Dataset and a Baseline",
    author = {H{\"a}m{\"a}l{\"a}inen, Mika  and
      Patpong, Pattama  and
      Alnajjar, Khalid  and
      Partanen, Niko  and
      Rueter, Jack},
    editor = "Xu, Wei  and
      Ritter, Alan  and
      Baldwin, Tim  and
      Rahimi, Afshin",
    booktitle = "Proceedings of W-NUT",
      year={2021}
}

@article{alghamdi2020predicting,
  title={Predicting depression symptoms in an Arabic psychological forum},
  author={Alghamdi, Norah Saleh and Mahmoud, Hanan A Hosni and Abraham, Ajith and Alanazi, Samar Awadh and Garc{\'\i}a-Hern{\'a}ndez, Laura},
  journal={IEEE access},
  volume={8},
  pages={57317--57334},
  year={2020},
  publisher={IEEE}
}

@article{figueroa2022automatic,
  title={Automatic language analysis identifies and predicts schizophrenia in first-episode of psychosis},
  author={Figueroa-Barra, Alicia and Del Aguila, Daniel and Cerda, Mauricio and Gaspar, Pablo A and Terissi, Lucas D and Dur{\'a}n, Manuel and Valderrama, Camila},
  journal={Schizophrenia},
  volume={8},
  number={1},
  pages={53},
  year={2022},
  publisher={Nature Publishing Group UK London}
}

@article{oh2024development,
  title={Development of depression detection algorithm using text scripts of routine psychiatric interview},
  author={Oh, Jihoon and Lee, Taekgyu and Chung, Eun Su and Kim, Hyonsoo and Cho, Kyongchul and Kim, Hyunkyu and Choi, Jihye and Sim, Hyeon-Hee and Lee, Jongseo and Choi, In Young and others},
  journal={Frontiers in psychiatry},
  volume={14},
  pages={1256571},
  year={2024},
  publisher={Frontiers Media SA}
}

@article{hull2020two,
  title={Two-way messaging therapy for depression and anxiety: longitudinal response trajectories},
  author={Hull, Thomas D and Malgaroli, Matteo and Connolly, Philippa S and Feuerstein, Seth and Simon, Naomi M},
  journal={BMC psychiatry},
  volume={20},
  number={1},
  pages={297},
  year={2020},
  publisher={Springer}
}

@article{marrie2018effects,
  title={Effects of psychiatric comorbidity in immune-mediated inflammatory disease: protocol for a prospective study},
  author={Marrie, Ruth Ann and Graff, Lesley and Walker, John R and Fisk, John D and Patten, Scott B and Hitchon, Carol A and Lix, Lisa M and Bolton, James and Sareen, Jitender and Katz, Alan and others},
  journal={JMIR Research Protocols},
  volume={7},
  number={1},
  pages={e8794},
  year={2018},
  publisher={JMIR Publications Inc., Toronto, Canada}
}

@article{rholes1985relationship,
  title={The relationship of cognitions and hopelessness to depression and anxiety},
  author={Rholes, William S and Riskind, John H and Neville, Brian},
  journal={Social Cognition},
  volume={3},
  number={1},
  pages={36--50},
  year={1985},
  publisher={Guilford Press}
}

@article{baryshnikov2020role,
  title={Role of hopelessness in suicidal ideation among patients with depressive disorders},
  author={Baryshnikov, Ilya and Rosenstr{\"o}m, Tom and Jylh{\"a}, Pekka and Vuorilehto, Maria and Holma, Mikael and Holma, Irina and Riihim{\"a}ki, Kirsi and Brown, Gregory K and Oquendo, Maria A and Isomets{\"a}, Erkki T},
  journal={The Journal of clinical psychiatry},
  volume={81},
  number={2},
  pages={8339},
  year={2020},
  publisher={Physicians Postgraduate Press, Inc.}
}

@article{page2021prisma,
  title={The PRISMA 2020 statement: an updated guideline for reporting systematic reviews},
  author={Page, Matthew J and McKenzie, Joanne E and Bossuyt, Patrick M and Boutron, Isabelle and Hoffmann, Tammy C and Mulrow, Cynthia D and Shamseer, Larissa and Tetzlaff, Jennifer M and Akl, Elie A and Brennan, Sue E and others},
  journal={bmj},
  volume={372},
  year={2021},
  publisher={British Medical Journal Publishing Group}
}

@article{chancellor2020methods,
  title={Methods in predictive techniques for mental health status on social media: a critical review},
  author={Chancellor, Stevie and De Choudhury, Munmun},
  journal={NPJ digital medicine},
  volume={3},
  number={1},
  pages={43},
  year={2020},
  publisher={Nature Publishing Group UK London}
}

@inproceedings{chancellor2019taxonomy,
  title={A taxonomy of ethical tensions in inferring mental health states from social media},
  author={Chancellor, Stevie and Birnbaum, Michael L and Caine, Eric D and Silenzio, Vincent MB and De Choudhury, Munmun},
  booktitle={Proceedings of FACCT},
  year={2019}
}

@article{chancellor2019human,
  title={Who is the" human" in human-centered machine learning: The case of predicting mental health from social media},
  author={Chancellor, Stevie and Baumer, Eric PS and De Choudhury, Munmun},
  journal={Proceedings of the ACM on Human-Computer Interaction},
  volume={3},
  number={CSCW},
  pages={1--32},
  year={2019},
  publisher={ACM New York, NY, USA}
}

@article{nicholas2020ethics,
  title={Ethics and privacy in social media research for mental health},
  author={Nicholas, Jennifer and Onie, Sandersan and Larsen, Mark E},
  journal={Current psychiatry reports},
  volume={22},
  pages={1--7},
  year={2020},
  publisher={Springer}
}

@article{matza2010identifying,
  title={Identifying HAM-A cutoffs for mild, moderate, and severe generalized anxiety disorder},
  author={Matza, Louis S and Morlock, Robert and Sexton, Chris and Malley, Karen and Feltner, Douglas},
  journal={International journal of methods in psychiatric research},
  volume={19},
  number={4},
  pages={223--232},
  year={2010},
  publisher={Wiley Online Library}
}

@article{demo1,
author = {Sunny Rai  and Elizabeth C. Stade  and Salvatore Giorgi  and Ashley Francisco  and Lyle H. Ungar  and Brenda Curtis  and Sharath C. Guntuku },
title = {Key language markers of depression on social media depend on race},
journal = {Proceedings of the National Academy of Sciences},
volume = {121},
number = {14},
pages = {e2319837121},
year = {2024}
}

@article{cao2024machine,
  title={Machine learning approaches for mental illness detection on social media: A systematic review of biases and methodological challenges},
  author={Cao, Yuchen and Dai, Jianglai and Wang, Zhongyan and Zhang, Yeyubei and Shen, Xiaorui and Liu, Yunchong and Tian, Yexin},
  journal={arXiv preprint arXiv:2410.16204},
  year={2024}
}

@article{uher2012self,
  title={Self-report and clinician-rated measures of depression severity: can one replace the other?},
  author={Uher, Rudolf and Perlis, Roy H and Placentino, Anna and Dernov{\v{s}}ek, Mojca Zvezdana and Henigsberg, Neven and Mors, Ole and Maier, Wolfgang and McGuffin, Peter and Farmer, Anne},
  journal={Depression and anxiety},
  volume={29},
  number={12},
  pages={1043--1049},
  year={2012},
  publisher={Wiley Online Library}
}

@article{arrow2023evaluating,
  title={Evaluating the use of online self-report questionnaires as clinically valid mental health monitoring tools in the clinical whitespace},
  author={Arrow, Kaitlyn and Resnik, Philip and Michel, Hanna and Kitchen, Christopher and Mo, Chen and Chen, Shuo and Espy-Wilson, Carol and Coppersmith, Glen and Frazier, Colin and Kelly, Deanna L},
  journal={Psychiatric Quarterly},
  volume={94},
  number={2},
  pages={221--231},
  year={2023},
  publisher={Springer}
}

@article{sylolypavan2023impact,
  title={The impact of inconsistent human annotations on AI driven clinical decision making},
  author={Sylolypavan, Aneeta and Sleeman, Derek and Wu, Honghan and Sim, Malcolm},
  journal={NPJ Digital Medicine},
  volume={6},
  number={1},
  pages={26},
  year={2023},
  publisher={Nature Publishing Group UK London}
}

@article{cruz2013appointment,
  title={Appointment length, psychiatrists’ communication behaviors, and medication management appointment adherence},
  author={Cruz, Mario and Roter, Debra L and Cruz, Robyn F and Wieland, Melissa and Larson, Susan and Cooper, Lisa A and Pincus, Harold Alan},
  journal={Psychiatric Services},
  volume={64},
  number={9},
  pages={886--892},
  year={2013},
  publisher={American Psychiatric Association Arlington, VA}
}

@inproceedings{coppersmith2017scalable,
  title={Scalable mental health analysis in the clinical whitespace via natural language processing},
  author={Coppersmith, Glen and Hilland, Casey and Frieder, Ophir and Leary, Ryan},
  booktitle={IEEE BHI},
  year={2017}
}

@article{stene2019contact,
  title={Contact with primary and mental health care prior to suicide: a systematic review of the literature from 2000 to 2017},
  author={Stene-Larsen, Kim and Reneflot, Anne},
  journal={Scandinavian journal of public health},
  volume={47},
  number={1},
  pages={9--17},
  year={2019},
  publisher={SAGE Publications Sage UK: London, England}
}

@article{althubaiti2016information,
  title={Information bias in health research: definition, pitfalls, and adjustment methods},
  author={Althubaiti, Alaa},
  journal={Journal of multidisciplinary healthcare},
  pages={211--217},
  year={2016},
  publisher={Taylor \& Francis}
}

@article{lynch2020substance,
  title={Substance use disorders and risk of suicide in a general US population: a case control study},
  author={Lynch, Frances L and Peterson, Edward L and Lu, Christine Y and Hu, Yong and Rossom, Rebecca C and Waitzfelder, Beth E and Owen-Smith, Ashli A and Hubley, Samuel and Prabhakar, Deepak and Keoki Williams, L and others},
  journal={Addiction science \& clinical practice},
  volume={15},
  number={1},
  pages={14},
  year={2020},
  publisher={Springer}
}

@article{bailey2011suicide,
  title={Suicide: current trends},
  author={Bailey, Rahn K and Patel, Tejas C and Avenido, Jaymie and Patel, Milapkumar and Jaleel, Mohammad and Barker, Narviar C and Khan, Jahanzeb Ali and All, Shahid and Jabeen, Shagufta},
  journal={Journal of the National Medical Association},
  volume={103},
  number={7},
  pages={614--617},
  year={2011},
  publisher={Elsevier}
}

@article{olteanu2019social,
  title={Social data: Biases, methodological pitfalls, and ethical boundaries},
  author={Olteanu, Alexandra and Castillo, Carlos and Diaz, Fernando and K{\i}c{\i}man, Emre},
  journal={Frontiers in big data},
  volume={2},
  pages={13},
  year={2019},
  publisher={Frontiers Media SA}
}

@article{zhao2022biases,
  title={Biases in using social media data for public health surveillance: A scoping review},
  author={Zhao, Yunpeng and He, Xing and Feng, Zheng and Bost, Sarah and Prosperi, Mattia and Wu, Yonghui and Guo, Yi and Bian, Jiang},
  journal={International Journal of Medical Informatics},
  volume={164},
  pages={104804},
  year={2022},
  publisher={Elsevier}
}

@article{hester2017lack,
  title={Lack of access to mental health services contributing to the high suicide rates among veterans},
  author={Hester, Ronald D},
  journal={International journal of mental health systems},
  volume={11},
  number={1},
  pages={47},
  year={2017},
  publisher={Springer}
}

@article{cummings2013improving,
  title={Improving access to mental health services for youth in the United States},
  author={Cummings, Janet R and Wen, Hefei and Druss, Benjamin G},
  journal={Jama},
  volume={309},
  number={6},
  pages={553--554},
  year={2013},
  publisher={American Medical Association}
}

@article{sametouglu2024value,
  title={The value of social media language for the assessment of wellbeing: A systematic review and meta-analysis},
  author={Sameto{\u{g}}lu, Selim and Pelt, DHM and Eichstaedt, Johannes C and Ungar, Lyle H and Bartels, Meike},
  journal={The Journal of Positive Psychology},
  volume={19},
  number={3},
  pages={471--489},
  year={2024},
  publisher={Taylor \& Francis}
}

@article{grant2009typology,
  title={A typology of reviews: an analysis of 14 review types and associated methodologies},
  author={Grant, Maria J and Booth, Andrew},
  journal={Health information \& libraries journal},
  volume={26},
  number={2},
  pages={91--108},
  year={2009},
  publisher={Wiley Online Library}
}

@article{zarate2022exploring,
  title={Exploring the digital footprint of depression: a PRISMA systematic literature review of the empirical evidence},
  author={Zarate, Daniel and Stavropoulos, Vasileios and Ball, Michelle and de Sena Collier, Gabriel and Jacobson, Nicholas C},
  journal={BMC psychiatry},
  volume={22},
  number={1},
  pages={421},
  year={2022},
  publisher={Springer}
}

@article{pazdur2025risk,
  title={Risk Factors for Problematic Social Media Use in Youth: A Systematic Review of Longitudinal Studies},
  author={Pazdur, Michelle and Tutus, Dunja and Haag, Ann-Christin},
  journal={Adolescent Research Review},
  pages={1--17},
  year={2025},
  publisher={Springer}
}

@article{templier2015framework,
  title={A framework for guiding and evaluating literature reviews},
  author={Templier, Mathieu and Par{\'e}, Guy},
  journal={Communications of the Association for Information Systems},
  volume={37},
  number={1},
  pages={6},
  year={2015}
}

@article{diederich2007ex,
  title={Ex-ray: Data mining and mental health},
  author={Diederich, Joachim and Al-Ajmi, Aqeel and Yellowlees, Peter},
  journal={Applied Soft Computing},
  volume={7},
  number={3},
  pages={923--928},
  year={2007},
  publisher={Elsevier}
}

@article{saini2014assessment,
  title={Assessment and management of suicide risk in primary care},
  author={Saini, Pooja and While, David and Chantler, Khatidja and Windfuhr, Kirsten and Kapur, Navneet},
  journal={Crisis},
  year={2014},
  publisher={Hogrefe Publishing}
}

@article{poulin2014predicting,
  title={Predicting the risk of suicide by analyzing the text of clinical notes},
  author={Poulin, Chris and Shiner, Brian and Thompson, Paul and Vepstas, Linas and Young-Xu, Yinong and Goertzel, Benjamin and Watts, Bradley and Flashman, Laura and McAllister, Thomas},
  journal={PloS one},
  volume={9},
  number={1},
  pages={e85733},
  year={2014},
  publisher={Public Library of Science San Francisco, USA}
}

@inproceedings{nobles2018identification,
  title={Identification of imminent suicide risk among young adults using text messages},
  author={Nobles, Alicia L and Glenn, Jeffrey J and Kowsari, Kamran and Teachman, Bethany A and Barnes, Laura E},
  booktitle={Proceedings of CHI},
  year={2018}
}

@article{shi2024enhancing,
  title={Enhancing depression diagnosis with chain-of-thought prompting},
  author={Shi, Elysia and Manda, Adithri and Chowdhury, London and Arun, Runeema and Zhu, Kevin and Lam, Michael},
  journal={arXiv preprint arXiv:2408.14053},
  year={2024}
}

@article{wei2022chain,
  title={Chain-of-thought prompting elicits reasoning in large language models},
  author={Wei, Jason and Wang, Xuezhi and Schuurmans, Dale and Bosma, Maarten and Xia, Fei and Chi, Ed and Le, Quoc V and Zhou, Denny and others},
  journal={Advances in neural information processing systems},
  volume={35},
  pages={24824--24837},
  year={2022}
}

@article{lee2023chain,
  title={Chain of empathy: Enhancing empathetic response of large language models based on psychotherapy models},
  author={Lee, Yoon Kyung and Lee, Inju and Shin, Minjung and Bae, Seoyeon and Hahn, Sowon},
  journal={arXiv preprint arXiv:2311.04915},
  year={2023}
}

@inproceedings{bucur2025survey,
  author={Bucur, Ana-Maria and Zampieri, Marcos and Ranasinghe, Tharindu and Crestani, Fabio},
  title={A Survey on Multilingual Mental Disorders Detection from Social Media Data},
  booktitle={Proceedings of EACL}, 
  year={2026},
}

@article{garg2023mental,
  title={Mental health analysis in social media posts: a survey},
  author={Garg, Muskan},
  journal={Archives of Computational Methods in Engineering},
  volume={30},
  number={3},
  pages={1819},
  year={2023}
}

@article{uzuner2017natural,
  title={A natural language processing challenge for clinical records: Research Domains Criteria (RDoC) for psychiatry},
  author={Uzuner, Ozlem and Stubbs, Amber and Filannino, Michele},
  journal={Journal of biomedical informatics},
  volume={75},
  pages={S1},
  year={2017}
}

@misc{GBD2021_IHME_2024,
  author       = {{Institute for Health Metrics and Evaluation}},
  title        = {Global Burden of Disease Study 2021 (GBD 2021) Results},
  year         = {2024},
  howpublished = {\url{https://vizhub.healthdata.org/gbd-results/}},
  note         = {Online database. Seattle, WA. Accessed 13 August 2025}
}

@misc{CDC_SuicidalThoughts_2025,
  author       = {{Centers for Disease Control and Prevention}},
  title        = {Suicidal Thoughts and Behavior},
  year         = {2025},
  howpublished = {\url{https://www.cdc.gov/mental-health/about-data/suicidal-thoughts-and-behavior.html}},
  note         = {Accessed 22 February 2026}
}

@inproceedings{gratch-distress,
    title = "The Distress Analysis Interview Corpus of human and computer interviews",
    author = "Gratch, Jonathan  and
      Artstein, Ron  and
      Lucas, Gale  and
      Stratou, Giota  and
      Scherer, Stefan  and
      Nazarian, Angela  and
      Wood, Rachel  and
      Boberg, Jill  and
      DeVault, David  and
      Marsella, Stacy  and
      Traum, David  and
      Rizzo, Skip  and
      Morency, Louis-Philippe",
    booktitle = "Proceedings of LREC",
    year = "2014"
}

@inproceedings{shin-dialogue,
    title = "Dialogue Summaries as Dialogue States ({DS}2), Template-Guided Summarization for Few-shot Dialogue State Tracking",
    author = "Shin, Jamin  and
      Yu, Hangyeol  and
      Moon, Hyeongdon  and
      Madotto, Andrea  and
      Park, Juneyoung",
    booktitle = "Findings of ACL",
    year = "2022"
}
\bibliographystyle{lrec2026-natbib}

\end{document}